\documentclass[10pt,twocolumn,letterpaper]{article}

 \usepackage[pagenumbers]{iccv} % To force page numbers, e.g. for an arXiv version

\usepackage[dvipsnames]{xcolor}

\usepackage{pifont}
\usepackage{fontawesome}
\usepackage{colortbl} 
\usepackage{times}
\usepackage{epsfig}
\usepackage{graphicx}
\usepackage{bm}
\usepackage{mathcommand}
\usepackage{mathrsfs}
\usepackage{amsmath,amsthm,amssymb,mathtools}
\usepackage{times}
\usepackage{epsfig}
\usepackage{graphicx}
\usepackage{amsmath}
\usepackage{amssymb}
\usepackage{bm}
\usepackage{amsthm}
\usepackage{algorithm}
\usepackage{algorithmicx}
\usepackage{algpseudocode}
\usepackage[dvipsnames]{xcolor}
\usepackage{caption}
\usepackage[export]{adjustbox}
\usepackage{float}
\usepackage{stfloats}
\usepackage{mathtools}
\usepackage{tabularx}
\usepackage{booktabs}
\usepackage{float}
\usepackage{placeins}
\usepackage{thm-restate}
\usepackage{thmtools}
\usepackage{wrapfig}
\usepackage{multirow}
\usepackage{enumitem}
\usepackage{colortbl}  
\usepackage{xcolor}
\usepackage{array}   
\usepackage{hhline} 
\usepackage{makecell}
\usepackage{wrapfig}
\usepackage{arydshln}
\newcommand{\bfe}{\mathbf{e}}

\definecolor{mypink1}{RGB}{241,90,82}
\definecolor{mypink}{RGB}{255,231,226}
\definecolor{myblue}{RGB}{233,250,249}

\definecolor{iccvblue}{rgb}{0.21,0.49,0.74}
\usepackage[pagebackref,breaklinks,colorlinks,allcolors=iccvblue]{hyperref}

\def\paperID{*****} % *** Enter the Paper ID here
\def\confName{ICCV}
\def\confYear{2025}

\title{Set You Straight: Auto-Steering Denoising Trajectories\\ to Sidestep Unwanted Concepts}

\author{
	{Leyang Li}$^{1,*}$
	\quad 
	{Shilin Lu}$^{1,*}$
	\quad 
	{Yan Ren}$^{1}$
	\quad 
	{Adams Wai-Kin Kong}$^{1}$\\
	\textsuperscript{1}Nanyang Technological University, Singapore \\
	{\tt\small \{lile0005, shilin002\}@e.ntu.edu.sg, nomatterhowlong@gmail.com, adamskong@ntu.edu.sg} \vspace{-0.4cm}
}

\begin{document}
\maketitle
\renewcommand{\thefootnote}{}
\footnotetext[1]{$*~$Equal contribution}
\begin{abstract}
	Ensuring the ethical deployment of text-to-image models requires effective techniques to prevent the generation of harmful or inappropriate content. While concept erasure methods offer a promising solution, existing finetuning-based approaches suffer from notable limitations. Anchor-free methods risk disrupting sampling trajectories, leading to visual artifacts, while anchor-based methods rely on the heuristic selection of anchor concepts. To overcome these shortcomings, we introduce a finetuning framework, dubbed \textbf{ANT}, which \textbf{A}utomatically guides de\textbf{N}oising \textbf{T}rajectories to avoid unwanted concepts. ANT is built on a key insight: reversing the condition direction of classifier-free guidance during mid-to-late denoising stages enables precise content modification without sacrificing early-stage structural integrity. This inspires a trajectory-aware objective that preserves the integrity of the early-stage score function field—which steers samples toward the natural image manifold—without relying on heuristic anchor concept selection. For single-concept erasure, we propose an augmentation-enhanced weight saliency map to precisely identify the critical parameters that most significantly contribute to the unwanted concept, enabling more thorough and efficient erasure. For multi-concept erasure, our objective function offers a versatile plug-and-play solution that significantly boosts performance. Extensive experiments demonstrate that ANT achieves state-of-the-art results in both single and multi-concept erasure, delivering high-quality, safe outputs without compromising the generative fidelity. Code is available at \href{https://github.com/lileyang1210/ANT}{https://github.com/lileyang1210/ANT}.
\end{abstract}

\section{Introduction}
Concept erasure in text-to-image (T2I) models~\cite{chang2023muse, ding2022cogview2, nichol2021glide, ramesh2022hierarchical, rombach2022high, saharia2022photorealistic, wang2024improving} addresses the critical challenge of preventing the generation of harmful or inappropriate visual content, such as violent, explicit, copyright-infringing, or offensive imagery. Current methods for concept erasure can be broadly categorized into two types: \textbf{(1)~finetuning-based methods}~\cite{lu2024mace,gandikota2023erasing,lyu2024one}, which directly modify model parameters, and \textbf{(2)~finetuning-free methods}~\cite{meng2025concept,schramowski2023safe,jain2024trasce}, which aim to influence model outputs without parameter updates. However, finetuning-free methods are vulnerable to bypassing when the source code is openly available, thus making finetuning-based methods more effective and secure for publicly accessible models.

Finetuning-based methods remove undesirable data modes by altering the predicted score function field—essentially, modifying the gradient directions that samples follow during the denoising process—to avoid converging toward undesirable image distributions. As a result of finetuning, the predicted score function no longer accurately reflects the true gradient direction in data space that would further increase likelihood. The main difference among finetuning-based techniques lies in how the conditional score function is modified, which can be broadly divided into \textbf{anchor-free} and \textbf{anchor-based} approaches.

\begin{figure*}[tbp]
	\centering
	%		\vspace{-0.2cm}
	\includegraphics[width=1\linewidth]{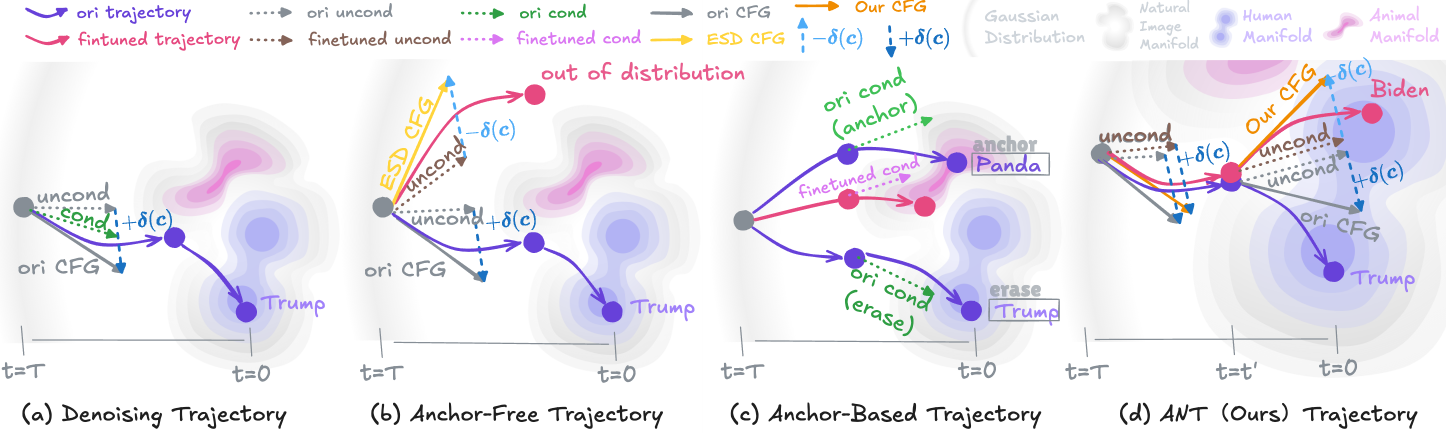}
%	\vspace{-0.6cm}
	\caption{
		Geometric perspective on concept erasure in diffusion models.
		\textbf{(a) Conventional Denoising Trajectory.} A high-dimensional Gaussian sample, starting on a large sphere, converges to the human data manifold via classifier-free guidance (CFG).
		\textbf{(b) Anchor-Free Finetuned Trajectory.} Finetuning often modifies the orientation of the predicted conditional score functions so that they direct away from the unwanted concept manifold. This results in a condition direction $\boldsymbol{\delta}(\boldsymbol{c}) = \boldsymbol{\epsilon}_{\boldsymbol{\theta}}(\boldsymbol{z}_{t}, t, \boldsymbol{c}) - \boldsymbol{\epsilon}_{\boldsymbol{\theta}}(\boldsymbol{z}_{t}, t)$ nearly opposite to that of the original model, making the trajectory more likely to produce out-of-distribution samples. Note that, in the absence of an unconditional constraint, modifications to the conditional output also affect the unconditional output due to shared model parameters. 
		\textbf{(c) Anchor-Based Finetuned Trajectory.} The model is finetuned so that the predicted score functions (or keys \& values) for the unwanted concept align with those of the original model conditioned on a benign anchor, ensuring final samples lie on the anchor manifold, though not necessarily at the highest-probability mode.
		\textbf{(d) Our Trajectory~(ANT).} In the early stage (when $t>t^\prime$), the conditional score functions remain directed toward the natural data mode, keeping the finetuned model aligned with the original. When $t<t^\prime$, they are finetuned to point away from the unwanted concept manifold. ANT encourages that unconditional score functions remain unchanged throughout all stages.}
	%		\vspace{-0.3cm}
	\label{fig:compare}
\end{figure*}

\begin{figure}[tbp]
	\centering
	%		\vspace{-0.2cm}
	\includegraphics[width=1\linewidth]{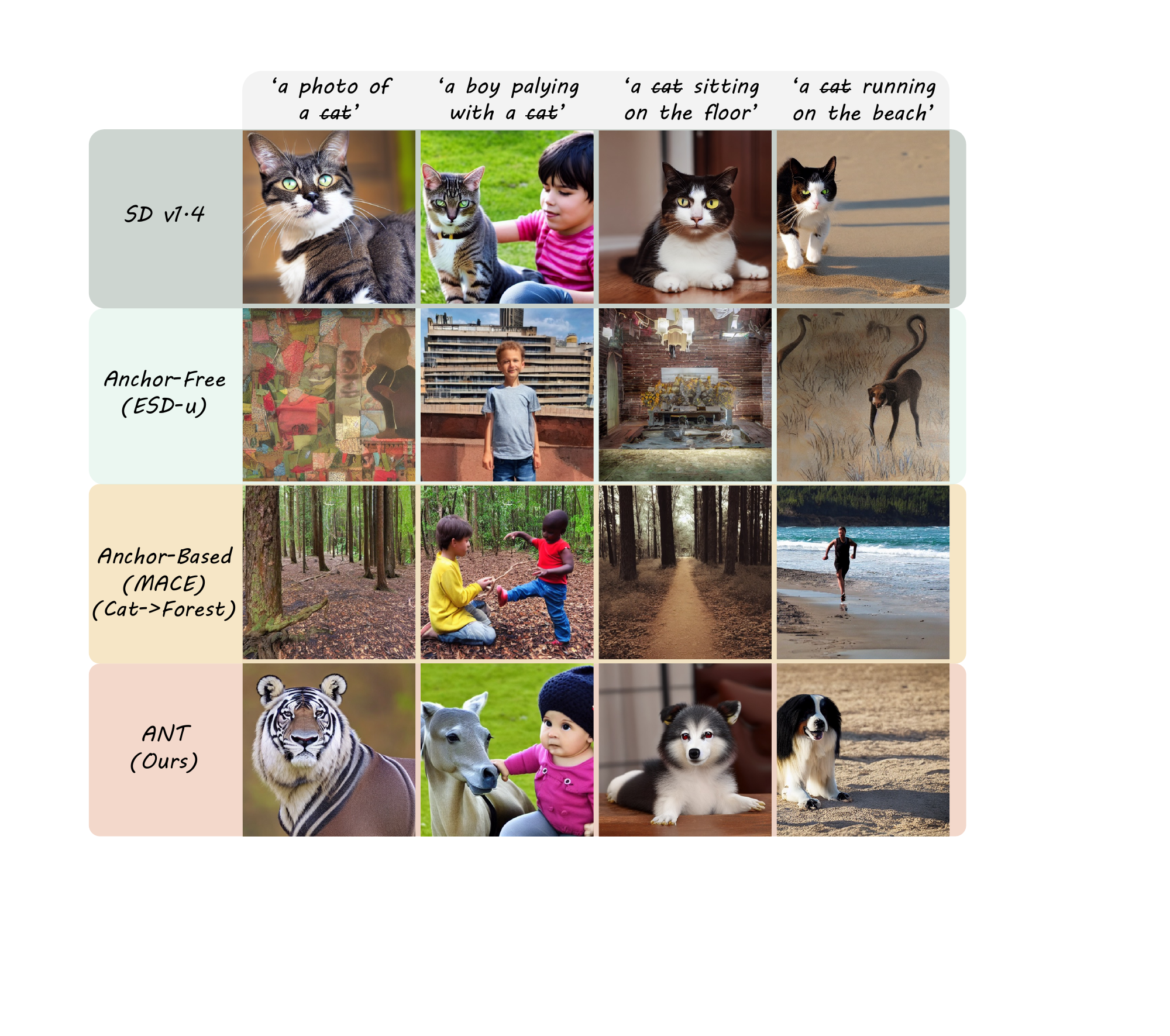}
%	\vspace{-0.6cm}
	\caption{Generation results of different concept erasure methods conditioned on the concept ``cat''. The anchor-free method (ESD) often produces images with visual artifacts or content that is out of distribution. The anchor-based method (MACE), which maps ``cat'' to ``forest'', performs reasonably well in simple contexts but results in unnatural or incoherent outputs in more complex scenarios. In contrast, our trajectory-aware method (ANT) effectively removes the target concept while preserving the overall structure and contextual integrity of the generated images.}
	\label{fig:failure}
	%		\vspace{-0.5cm}
\end{figure}

%\vspace{0.2cm}
\textbf{Anchor-free methods}~\cite{lyu2024one,wang2025ace,gandikota2023erasing,wang2025ace2,schramowski2023safe,kim2023towards,kim2024safeguard,nguyen2024unveiling,zhang2024defensive,heng2023selective,han2025dumo,wu2024munba,liu2024implicit,thakral2025fine,gao2024eraseanything,kim2024safety,huang2024receler,hong2024all,bui2024erasing,kim2024race,yang2024pruning,chengrowth,schioppa2024model,tu2025sdwv,cywinski2025saeuron,meng2025concept,thakral2025continual,chen2025safe} often design a loss to adjust the conditional score function throughout the denoising process, encouraging samples to move away from unwanted image manifolds without explicitly specifying a target manifold (see Figure~\ref{fig:compare}(b)). However, this approach can disrupt the sampling trajectories toward natural image manifolds. As shown in Figure~\ref{fig:compare}(a), diffusion models typically first guide samples from Gaussian noise toward the manifold of natural images to establish a plausible layout, and then progressively refine the details during the mid-to-late denoising steps~\cite{kwon2022diffusion,lu2024mace}. By solely emphasizing the movement away from unwanted manifolds, anchor-free methods risk causing samples to deviate from the natural image manifold early on, potentially resulting in generated images with visual artifacts or unintended content (see the second row of Figure~\ref{fig:failure}).
%	Our preliminary experiments indicate that even when the ESD loss is confined solely to the mid-to-late denoising steps—intended to preserve the early-stage score function field—the effect on the early-stage score field persists, likely due to the model parameters being shared across all timesteps in the diffusion process.

\textbf{Anchor-based methods}~\cite{xiong2024editing,srivatsan2024stereo,kumari2023ablating,gong2024reliable,bui2024erasing,bui2025fantastic,beerens2025vulnerability,chen2025trce,li2025speed,kim2025concept,zhang2023forget,chavhan2024conceptprune,liu2024realera,lu2024mace,lee2025localized,han2024continuous,park2024direct,fuchi2024erasing,shirkavand2024efficient,gandikota2024unified,zhao2024advanchor,fuchi2025erasing,hu2025safetext,tian2025sparse,xue2025crce,li2025detect}, on the other hand, typically utilize a loss designed to leverage benign anchor concepts by aligning the predicted conditional score functions (or keys \& values) for unwanted concepts with those associated with anchor concepts (see Figure~\ref{fig:compare}(c)). By aligning score functions of unwanted concepts with those of anchor concepts, these methods ensure that samples conditioned on unwanted concepts ultimately converge towards images depicting the anchor concepts. Thus, anchor-based approaches are not merely repelling samples from undesired modes. Nevertheless, the effectiveness of these methods critically depends on the proper selection of anchor concepts. As demonstrated in the third row of Figure~\ref{fig:failure}, some seemingly reasonable anchor concept choices can reduce the quality of images generated when conditioned on erased concepts. Currently, selecting effective anchor concepts remains largely heuristic, lacking systematic guidelines. 

Motivated by these limitations, we propose a trajectory-aware finetuning framework, termed \textbf{ANT}, which \textbf{A}utomatically guides de\textbf{N}oising \textbf{T}rajectories to avoid unwanted concepts. This approach achieves its goal without negatively affecting early-stage score function fields or relying on heuristic anchor concept selection. Specifically, we discovered that reversing the condition direction of classifier-free guidance (CFG)~\cite{ho2022classifier} during the mid-to-late denoising stage enables modification of detailed content while preserving the fundamental structure of the generated image. This finding inspires us to develop a trajectory-aware objective function that preserves the early-stage score function, steering samples toward the natural image manifold, while eliminating the need for anchor concepts (Figure~\ref{fig:compare}(d)). This approach enables more effective erasure of undesired concepts while better preserving those that are unrelated. In the context of single-concept erasure, we introduce an augmentation-enhanced weight saliency map that accurately identifies the key parameters most responsible for generating a specific concept. Moreover, our loss function is fully compatible with existing multi-concept erasure frameworks, offering a flexible plug-and-play solution, and elevates the performance to a new state-of-the-art (SOTA) level. Our experimental results demonstrate that our method achieves SOTA performance in both single and multi-concept erasure settings. Our contributions are summarized as follows:

\begin{itemize}
		\setlength{\itemsep}{4pt}
		\setlength{\parsep}{0pt}
		\setlength{\parskip}{0pt}
		\item [1.] We offer a geometric perspective on concept erasure and an insight that reversing the condition direction of classifier-free guidance during the mid-to-late denoising stages enables precise content modification while preserving early-stage structural integrity, thus benefiting the erasure community in advancing algorithm designs.
		\item [2.] We propose a trajectory-aware finetuning framework, which encourages the model to reorient its denoising trajectories during the mid-to-late stages while keeping the early-stage trajectories largely unchanged. This approach enables more thorough erasure of unwanted concepts and better preservation of unrelated ones.
		\item [3.] We introduce an augmentation-enhanced weight saliency map that precisely identifies the key parameters most responsible for generating the undesired concept, thereby enabling more effective and efficient erasure.
		\item [4.] The proposed objective function substantially enhances the performance of existing multi-concept erasure frameworks, achieving SOTA results in both single- and multi-concept erasure settings.  
\end{itemize}

\section{Related Work}
In this section, we review prior work on concept erasure in diffusion models, with a particular focus on the critical trade-off between erasure and preservation, which is most pertinent to our study. Additional discussions on other dimensions of concept erasure (e.g., finetuning efficiency, scalability, and robustness to adversarial prompts) are provided in Appendix.

The investigation of concept erasure within diffusion models has been pioneered by several foundational studies, establishing the groundwork for this burgeoning domain. SLD \cite{schramowski2023safe} introduces an inference-time guidance technique to suppress undesired concepts without modifying the model’s parameters, offering a non-invasive yet effective approach. In contrast, ESD \cite{gandikota2023erasing} employs direct parameter editing through negative guidance, achieving permanent concept removal. FMN \cite{zhang2023forget} builds upon this trajectory by proposing a lightweight method that manipulates attention mechanisms to enhance computational efficiency. Meanwhile, AC \cite{kumari2023ablating} presents a finetuning framework that aligns the score function of an unwanted concept with that of an anchor concept, delivering an alternative strategy for concept ablation.

As concept erasure techniques have matured, the research community has increasingly emphasized the dual objectives of effectively eliminating target concepts while preserving the integrity of unrelated concepts during the finetuning process. Numerous studies~\cite{lyu2024one,wang2025ace,wang2025ace2,kim2023towards,kim2024safeguard,bui2024erasing,bui2025fantastic,gaintseva2025casteer,gao2024meta,chavhan2024conceptprune,heng2023selective,han2025dumo,wu2024munba,liu2024realera,lu2024mace,thakral2025fine,han2024continuous,huang2024receler,yang2024pruning,shirkavand2024efficient,chengrowth,schioppa2024model,zhao2024advanchor,tu2025sdwv,fuchi2025erasing,meng2025concept,tian2025sparse,xue2025crce,thakral2025continual,li2025detect,chen2025safe,gao2024eraseanything} highlight the necessity of maintaining balanced model performance across both targeted and non-targeted concepts. However, a critical limitation of these approaches lies in their insufficient attention to the impacts of finetuning on the early-stage score function. This oversight can lead to a divergence between the predicted score function and the true score function, i.e., the gradient direction in data space that maximizes likelihood. As a result, the generated samples may fail to converge toward the natural image manifold, ultimately degrading the quality and reliability of the outputs. Our work seeks to bridge this gap by explicitly addressing the preservation of the early-stage score function, ensuring both effective concept erasure and high-fidelity generation.

\section{Method}
We propose ANT, a framework designed to erase specific concepts from pretrained text-to-image diffusion models. Our approach addresses key challenges by eliminating the negative impacts on early-stage score function fields and removing the dependency on heuristic methods for anchor concept selection. The framework requires only two inputs: a pretrained diffusion model and a set of target phrases representing the concepts to be erased. The output is a finetuned model that no longer generates images depicting the unwanted concepts.

\subsection{Insights into the Denoising Process}

We thoroughly investigated the denoising process in diffusion models and found that applying CFG during the early sampling stage (when $t^\prime < t < T$), and then reversing the CFG's condition direction term during the mid-to-late sampling stage (when $0<t<t^\prime$, as shown in Eq.~(\ref{eq:cfg})), allows for altering detailed content while preserving the fundamental structure of the image. In other words, the sample avoids converging toward specific unwanted concepts yet remains within the natural image manifold.
\begin{align} 
	\boldsymbol{\epsilon}^\text{cfg}_{\boldsymbol{\theta}}(\boldsymbol{z}_{t}, t, \boldsymbol{c}) = \boldsymbol{\epsilon}_{\boldsymbol{\theta}}(\boldsymbol{z}_{t}&, t) + s \cdot \operatorname{sgn}(t - t^\prime) \cdot \boldsymbol{\delta}(\boldsymbol{c}), \label{eq:cfg} \\ 
	\operatorname{sgn}(t-t^\prime) &= 
	\begin{cases}
		-1, & \text{if } t \le t^\prime \\
		1,  & \text{if } t > t^\prime
	\end{cases}
\end{align} 
where the terms $\boldsymbol{\epsilon}^\text{cfg}_{\boldsymbol{\theta}}(\boldsymbol{z}{t}, t, \boldsymbol{c})$, $\boldsymbol{\epsilon}_{\boldsymbol{\theta}}(\boldsymbol{z}{t}, t, \boldsymbol{c})$, and $\boldsymbol{\epsilon}_{\boldsymbol{\theta}}(\boldsymbol{z}_{t}, t)$ denote the classifier-free guidance output, the conditional prediction, and the unconditional prediction, respectively. The difference $\boldsymbol{\delta}(\boldsymbol{c}) = \boldsymbol{\epsilon}_{\boldsymbol{\theta}}(\boldsymbol{z}_{t}, t, \boldsymbol{c}) - \boldsymbol{\epsilon}_{\boldsymbol{\theta}}(\boldsymbol{z}_{t}, t)$ defines the condition direction. $t^\prime$ is a key parameter used to determine the timestep at which the condition direction should be reversed.

\begin{figure}[tbp]
	\centering
	%		\vspace{-0.2cm}
	\includegraphics[width=1\linewidth]{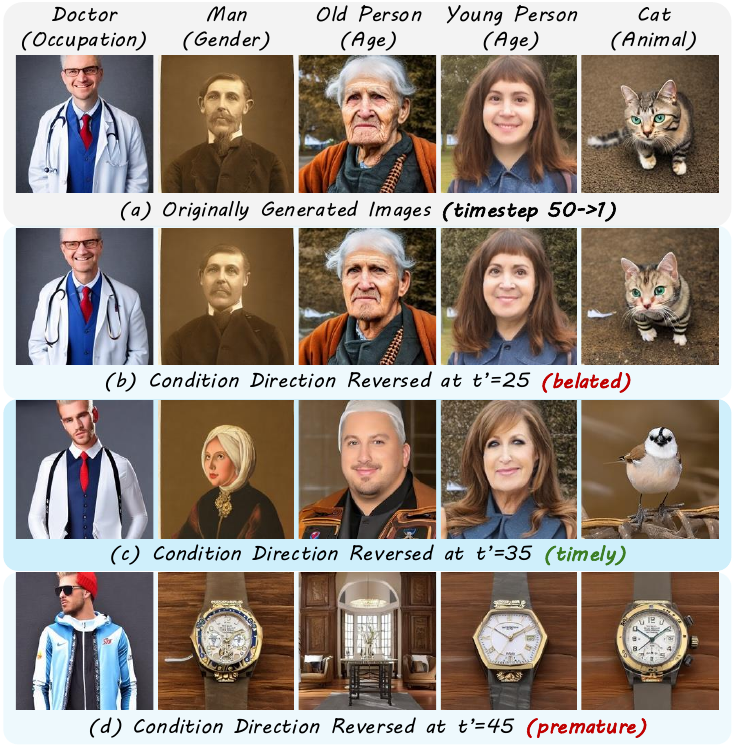}
%	\vspace{-0.7cm}
	\caption{Effect of condition direction reversal at different timesteps. Each column represents a distinct semantic condition, and each row shows generated outputs under varying reversal strategies. (a) displays originally generated images using a diffusion process (timestep 50→1). (b)–(d) show results when the condition direction $\boldsymbol{\delta}(\boldsymbol{c}) = \boldsymbol{\epsilon}_{\boldsymbol{\theta}}(\boldsymbol{z}_{t}, t, \boldsymbol{c}) - \boldsymbol{\epsilon}_{\boldsymbol{\theta}}(\boldsymbol{z}_{t}, t)$ is reversed at different timesteps (25, 35, and 45). With a proper \( t' \), specific attributes can be removed while preserving image naturalness. If \( t' \) is too early, structural integrity is lost; if too late, only fine details are affected.}
	\label{fig:cfg}
	%		\vspace{-0.3cm}
\end{figure}

As shown in Figure~\ref{fig:cfg}(c), if $t^\prime$ is appropriately selected, this approach allows for the targeted removal of specific attributes or details (e.g., occupation, gender, or age) while preserving the naturalness of the generated images. This is because, during the early stage of denoising, the samples follow the correct score function and are guided onto a plausible data manifold. In the later stages, the guidance steers the samples away from certain modes within that manifold. For instance, in Figure~\ref{fig:cfg}(c), the occupation changes from doctor to model, gender shifts from male to female, and age transitions from both old and young to middle-aged, all while staying within the human data manifold. 

However, if \( t' \) is set too early, the early-stage score function will be significantly altered, leading to a loss of the image's structural integrity (see Figure~\ref{fig:cfg}(d)). On the other hand, if \( t' \) is set too late, the samples will have already entered the concept-specific mode, and modifications to the late-stage score function will only affect fine details (see Figure~\ref{fig:cfg}(b)).

\subsection{Trajectory-Aware Loss Function}
Inspired by this finding, we aim to preserve the integrity of the early-stage score function field—which guides samples toward the appropriate natural manifold—by introducing constraints during finetuning. Adjustments will be limited exclusively to the mid-to-late stage score function field. This approach ensures that even when the finetuned model is conditioned on the removed concept, the samples can still converge to the appropriate manifold. Specifically, we propose the following finetuning objective:
\begin{equation}
	\resizebox{\columnwidth}{!}{$
		\begin{aligned}
			\mathcal{L} & = \mathcal{L}_\text{preserve} + \lambda_1 \cdot \mathcal{L}_\text{erase} + \lambda_2 \cdot \mathcal{L}_\text{uncond-early} + \lambda_3 \cdot \mathcal{L}_\text{uncond-late}  \\  
			& = \mathbb{E}_{\boldsymbol{z}_{t_1},\boldsymbol{c}, t_1 \sim U(t^\prime, T)} \left[ \left\| \boldsymbol{\epsilon}_{\boldsymbol{\theta}}(\boldsymbol{z}_{t_1}, t_1, \boldsymbol{c}) - {\tt sg} \left[ \boldsymbol{\epsilon}_{\boldsymbol{\theta}^*}(\boldsymbol{z}_{t_1}, t_1) + \eta \boldsymbol{\delta}(\boldsymbol{c}) \right]  \right\|^2_2 \right]  \\
			& + \lambda_1 \mathbb{E}_{\boldsymbol{z}_{t_2},\boldsymbol{c}, t_2 \sim U(0, t^\prime)} \left[ \left\| \boldsymbol{\epsilon}_{\boldsymbol{\theta}}(\boldsymbol{z}_{t_2}, t_2, \boldsymbol{c}) - {\tt sg} \left[ \boldsymbol{\epsilon}_{\boldsymbol{\theta}^*}(\boldsymbol{z}_{t_2}, t_2) - \eta \boldsymbol{\delta}(\boldsymbol{c}) \right]  \right\|^2_2 \right]  \\
			& + \lambda_2 \mathbb{E}_{\boldsymbol{z}_{t_1},\boldsymbol{c}, t_1 \sim U(t^\prime, T)} \left[ \left\| \boldsymbol{\epsilon}_{\boldsymbol{\theta}}(\boldsymbol{z}_{t_1}, t_1) - {\tt sg} \left[ \boldsymbol{\epsilon}_{\boldsymbol{\theta}^*}(\boldsymbol{z}_{t_1}, t_1) \right] \right\|^2_2 \right]  \\
			& + \lambda_3 \mathbb{E}_{\boldsymbol{z}_{t_2},\boldsymbol{c}, t_2 \sim U(0, t^\prime)} \left[ \left\| \boldsymbol{\epsilon}_{\boldsymbol{\theta}}(\boldsymbol{z}_{t_2}, t_2) - {\tt sg} \left[ \boldsymbol{\epsilon}_{\boldsymbol{\theta}^*}(\boldsymbol{z}_{t_2}, t_2) \right] \right\|^2_2 \right],
		\end{aligned}
		$}
	\label{eq:loss}
\end{equation}
where $\boldsymbol{\theta}$ represents the parameters undergoing finetuning, while $\boldsymbol{\theta}^*$ denotes the original, frozen parameters. The notation ${\tt sg}[\cdot]$ indicates the stop-gradient operation. Timesteps $t_1$ and $t_2$ are sampled independently from uniform distributions $U(t^\prime, T)$ and $U(0, t^\prime)$, respectively, with $t^\prime$ being a predefined hyperparameter. Additionally, $\boldsymbol{z}_{t_1}$ and $\boldsymbol{z}_{t_2}$ represent the corresponding noisy latent image variables at these timesteps, and $\eta$ denotes a hyperparameter. \textit{Notably, two timesteps are sampled during each gradient update iteration to effectively balance the gradients associated with concept erasure and the preservation of unrelated concepts.}

\vspace{0.1cm}
\noindent \textbf{Early-stage preservation.} The first term $\mathcal{L}_\text{preserve}$ ensures that, during the early stage (when $t > t'$), the predicted conditional score function consistently points toward the natural data mode. This preserves the integrity of the early stage score function field. Consequently, when sampling with the finetuned model conditioned on the erased concept, the generated samples can smoothly transition into the natural image manifold. 

\vspace{0.1cm}
\noindent \textbf{Mid-to-late-stage erasure.} The second term $\mathcal{L}_\text{erase}$ emphasizes that at later stage (when $t < t'$), the predicted conditional score function should actively guide samples away from undesirable modes. It differs from the ESD loss~\cite{gandikota2023erasing} in that the second term is applied exclusively at later timesteps ($t < t'$), whereas the ESD loss spans all timesteps. Including early timesteps in the ESD loss can unintentionally alter the early-stage score function field, frequently causing samples to be incorrectly guided and thereby failing to converge onto the appropriate manifold. To further explore this issue, we conducted an experiment restricting the application of this second loss term solely to mid-to-late denoising steps, specifically aiming to avoid negatively impacting the early-stage score function field. However, even under this restricted condition, the early-stage score function field was still adversely affected, resulting in suboptimal performance (see the ablation study in Table 2). We hypothesize that this outcome arises primarily due to the shared model parameters across all timesteps within the diffusion process.

\vspace{0.1cm}
\noindent \textbf{Unconditional score function preservation.} Since the unconditional score function $\boldsymbol{\epsilon}_{\boldsymbol{\theta}}(\boldsymbol{z}_{t}, t)$ represents the general direction toward the approximate center of all data modes, modifying it can influence multiple concepts, as demonstrated by our ablation study. Specifically, Table 2 shows that removing 100 celebrity concepts without incorporating unconditional loss terms negatively impacts the preservation of other celebrity concepts. To address this issue, we introduce the third and fourth terms in Eq.~(\ref{eq:loss}). These terms align the unconditional outputs of the finetuned model with those of the original model across both stages. 

\subsection{The Heavy Hitters Among the Parameters}

After determining the optimization objective, identifying the most effective parameters to optimize for achieving improved performance efficiently becomes crucial. Previous approaches typically divide the model into multiple modules, such as residual blocks, self-attention, or cross-attention, and select an entire module for finetuning~\cite{gandikota2023erasing,gandikota2024unified,lu2024mace,zhang2023forget}. Among these, finetuning cross-attention modules is most common. 

\begin{figure}[tbp]
	\centering
	%		\vspace{-0.2cm}
	\includegraphics[width=1\linewidth]{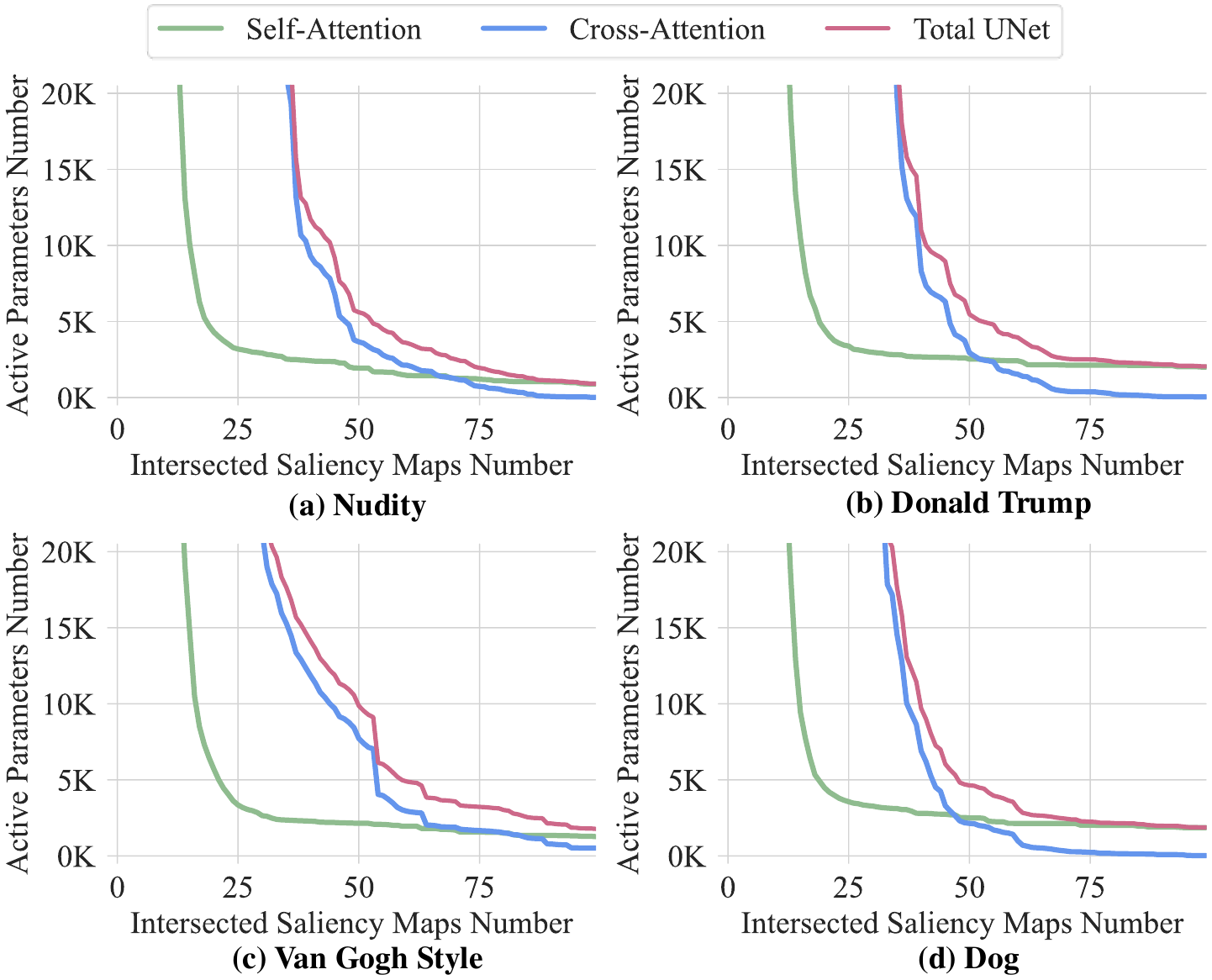}
%	\vspace{-0.5cm}
	\caption{Each subplot shows the number of active parameters (y-axis) against the number of intersected saliency maps (x-axis) for four concepts: (a) Nudity, (b) Donald Trump, (c) Van Gogh Style, and (d) Dog. The number of active parameters converges across different concept types with around 100 intersected saliency maps.}
	\label{fig:saliency_1}
	%		\vspace{-0.3cm}
\end{figure}

\begin{figure}[tbp]
	\centering
	%		\vspace{-0.2cm}
	\includegraphics[width=1\linewidth]{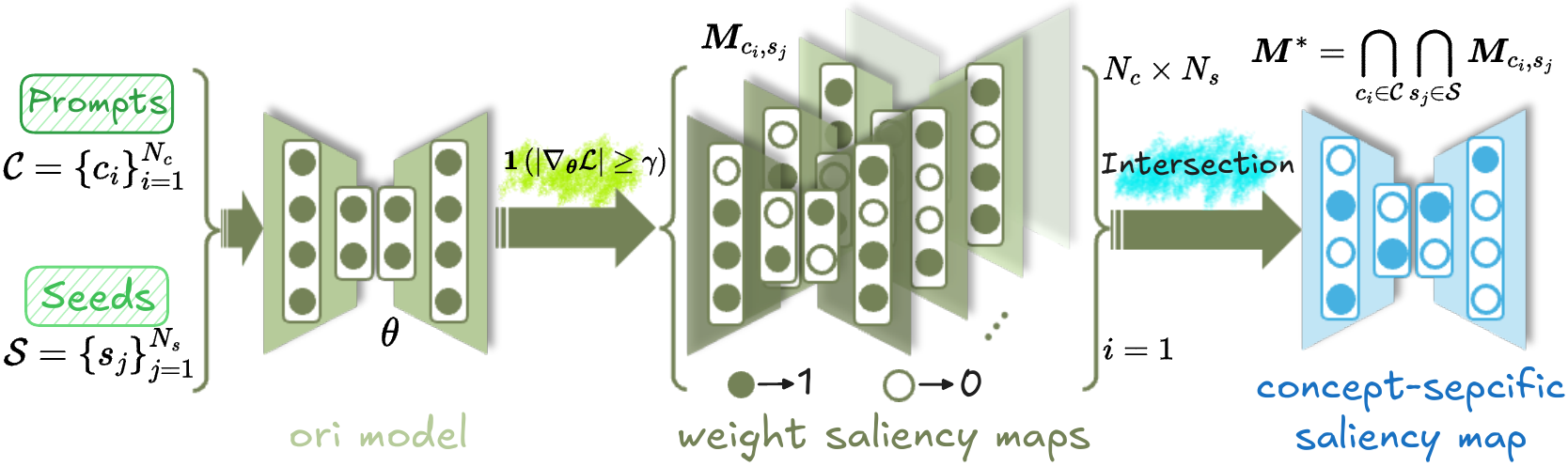}
%	\vspace{-0.7cm}
	\caption{Generation of the concept-specific saliency map \(\boldsymbol{M}^*\). GPT-4 generates prompts \(\mathcal{C} = \{c_i\}_{i=1}^{N_c}\), each paired with random seeds \(\mathcal{S} = \{s_j\}_{j=1}^{N_s}\), which are used to compute gradient maps. After thresholding, saliency maps are obtained, and their intersection across all prompts and seeds yields \(\boldsymbol{M}^*\).}
	\label{fig:saliency}
	%		\vspace{-0.3cm}
\end{figure}

Inspired by saliency map techniques~\cite{fan2023salun,simonyan2013deep,smilkov2017smoothgrad,sundararajan2017axiomatic,han2015deep,frankle2018lottery}, we propose a concept-specific saliency map enhanced by prompt and seed augmentation to precisely identify parameters suitable for finetuning. Compared to previous methods that compute the saliency map only once, we observe that the saliency map can vary depending on the prompt context and random seed, leading to instability. However, if we take the intersection of multiple saliency maps, the parameters within this intersection gradually become more stable and consistent as the number of maps increases (see Figure~\ref{fig:saliency_1}). This approach more accurately identifies the parameters responsible for the target concept, resulting in a consistent improvement in performance (as shown in the ablation study results in Table~\ref{tab:ablation_nsfw}). Specifically, as illustrated in Figure~\ref{fig:saliency}, we first employ GPT-4~\cite{GPT-4o} to generate multiple prompts \( \mathcal{C} = \{c_i\}_{i=1}^{N_c} \), each accompanied by a set of random seeds \(\mathcal{S} = \{s_j\}_{j=1}^{N_s} \), to produce corresponding gradient maps for the model parameters. By evaluating these gradients against a threshold, we obtain a set of weight saliency maps:
\begin{align}
	\boldsymbol{M}_{c_i, s_j} = \mathbf{1}\left( |\nabla_{\boldsymbol{\theta}}\mathcal{L}(\boldsymbol{z}_{t_1}, \boldsymbol{z}_{t_2}, t_1, t_2, c_i, s_j)| \geq \gamma \right),
\end{align}
where $\mathbf{1}(\boldsymbol{g} \geq \gamma)$ is an element-wise indicator function that returns 1 for the $i$-th element if $g_i \geq \gamma$, and 0 otherwise; $|\cdot|$ denotes the element-wise absolute value operation; and $\gamma > 0$ is a predefined threshold. Each weight saliency map identifies critical parameters strongly correlated with the targeted concept across diverse prompt contexts. Finally, the intersection of these weight saliency maps obtained from various prompts and seeds yields the definitive concept-specific saliency map $\boldsymbol{M}^*$:
\begin{equation}
	\boldsymbol{M}^* = \bigcap_{c_i \in \mathcal{C}} \bigcap_{s_j \in \mathcal{S}} \boldsymbol{M}_{c_i, s_j}.
\end{equation}
As a result, only a crucial subset of parameters is finetuned:
\begin{equation}
	\boldsymbol{\theta} \leftarrow \boldsymbol{\theta} - \alpha \cdot \boldsymbol{M}^* \odot \nabla_{\boldsymbol{\theta}}\mathcal{L}(\boldsymbol{z}_{t_1}, \boldsymbol{z}_{t_2}, t_1, t_2, c_i, s_j),
\end{equation}
where $\alpha$ is the learning rate and $\odot$ denotes the element-wise multiplication. Intuitively, this mechanism identifies and finetunes only those parameters consistently influential for erasing the undesired concept across diverse conditions. Concept-specific saliency map~$\boldsymbol{M}^*$ significantly narrows down the finetuning parameters, effectively preventing unnecessary perturbations to parameters unrelated to the targeted concept.

\subsection{Boosting the Performance of Multi-Concept Erasure Frameworks}
Our proposed trajectory-aware loss function seamlessly integrates with existing multi-concept erasure frameworks, such as MACE~\cite{lu2024mace}, offering a flexible and adaptable plug-and-play solution. Accordingly, it significantly boosts MACE’s performance in multi-concept scenarios, delivering new SOTA outcomes on tasks involving the erasure of 100 celebrity concepts and 100 artistic concepts. 

As observed in MACE, erasing multiple concepts through either sequential or parallel finetuning often degrades performance. Sequential finetuning is susceptible to catastrophic forgetting, while parallel finetuning can lead to interference between concepts  \cite{lu2024mace}. MACE addresses this by training a separate LoRA module for each concept to be erased, and subsequently fusing all LoRA modules into the cross-attention layers using a closed-form solution.

By integrating our loss function into the MACE framework, the initial training stage can be omitted. In the second stage, we replace MACE’s attention loss with our trajectory-aware loss to train individual LoRA modules $\Delta \boldsymbol{W}_i$ for each concept, eliminating the need for the large Grounded-SAM model. After training all LoRA modules, we use the following objective function to fuse them into the cross-attention layers:
\begin{equation} 
	\label{eq:close-form-fuse}
	\begin{split}
		\min\limits_{\boldsymbol{W}^*} &\sum\limits_{i=1}^q \sum\limits_{j=1}^p \left\|  \boldsymbol{W}^* \cdot \bfe_j^f  - \left(\boldsymbol{W} + \Delta \boldsymbol{W}_{i}\right) \cdot \bfe^f_j \right\|_2^2 \\
		& +  \beta \sum\limits_{j=p+1}^{p+m} \left\| \boldsymbol{W}^* \cdot \bfe_j^p - \boldsymbol{W} \cdot \bfe^p_j  \right\|_2^2,
	\end{split}
\end{equation}
where $\boldsymbol{W}$ denotes the original weight matrix of either the key or value projection. The embedding $e_j^f$ corresponds to concept-related tokens that we aim to erase, while $e_j^p$ represents embeddings of unrelated, prior-preservation tokens. Here, $q$ is the number of concepts to be erased, and $p$ and $m$ denote the numbers of targeted concept tokens and prior-preservation tokens, respectively. 

\begin{figure}[tbp]
	\centering
	%		\vspace{-0.2cm}
	\includegraphics[width=1\linewidth]{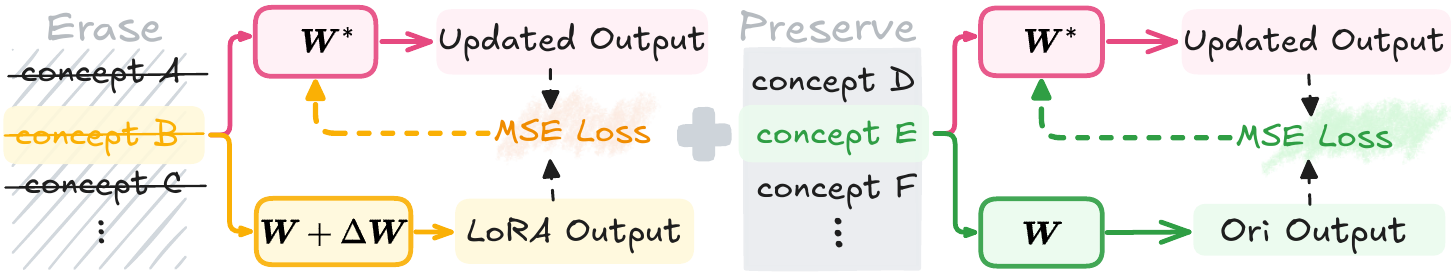}
%	\vspace{-0.7cm}
	\caption{Multi-LoRA fusion for multi-concept erasure.}
	\label{fig:lora}
%	\vspace{-0.3cm}
\end{figure}

As shown in Figure~\ref{fig:lora}, the objective is to find a solution $\boldsymbol{W}^*$ that integrates multiple LoRA matrices, optimized for effective multi-concept erasure. This optimization problem has a closed-form solution~\cite{lu2024mace}. Table~\ref{tab:celebrity} shows that our trajectory-aware loss function seamlessly integrates with the MACE framework for multi-concept erasure, substantially enhancing its performance.

\section{Experiments}

\begin{table*}[tbp]
	\centering
	\caption{Results of Erasing NSFW Content. The left side shows the number of exposed body parts detected on the I2P dataset using the NudeNet detector, while the right side presents the FID and CLIP on the COCO dataset. M:~Male. F:~Female. }
	%	\vspace{-0.3cm}
	\label{tab:nudenet-results}
	\resizebox{\textwidth}{!}{
		\begin{tabular}{lcccccccc>{\columncolor{mypink}}ccc}
			\toprule
			\multirow{2}{*}{{Method}} 
			& \multicolumn{9}{c}{{Inappropriate Image Prompt (I2P)}} 
			& \multicolumn{2}{c}{{MS-COCO 30K}} \\
			\cmidrule(lr){2-10} \cmidrule(lr){11-12}
			& {Armpits} & {Belly} & {Buttocks} & {Feet} 
			& {Breasts (F)} & {Genitalia (F)} 
			& {Breasts (M)} & {Genitalia (M)} 
			& {Total $\downarrow$} 
			& {FID $\downarrow$} & {CLIP $\uparrow$} \\
			\midrule
			FMN~\cite{zhang2023forget}  &  43 & 117 & 12 & 59 & 155 & 17 & 19 & 2 & 424 & 13.52 & 30.39 \\
			ESD-x~\cite{gandikota2023erasing}&  59 &  73 & 12 & 39 & 100 &  6 & 18 & 8 & 315 & 14.41 & 30.69 \\
			ESD-u~\cite{gandikota2023erasing}&  32 &  30 &  2 & 19 &  27 &  3 &  8 & 2 & 123 & 15.10 & 30.21 \\
			SLD-M~\cite{schramowski2023safe}&  47 &  72 &  3 & 21 &  39 &  {1} & 26 & 3 & 212 & 16.34 & 30.90 \\
			AC~\cite{kumari2023ablating}   & 153 & 180 & 45 & 66 & 298 & 22 & 67 & 7 & 838 & 14.13 & \textbf{31.37} \\
			SA~\cite{heng2023selective} 	&  72 &  77 & 19 & 25 &  83 & 16 &  {0} & {0} & 292 & --    & --    \\
			EA \cite{gao2024eraseanything} & - & - & - & - & - & - & - & - & 199 & 21.75 & 30.24 \\
			UCE~\cite{gandikota2024unified}  &  29 &  62 &  7 & 29 &  35 &  5 & 11 & 4 & 182 & 14.07 & 30.85 \\
			Receler\cite{huang2024receler} & 39 & 26 & 5 & 10 & 13 & 1 & 12 & 9 & 115 & - & -\\
			MACE~\cite{lu2024mace} & 17 & 19 & 2 & 39 & 16 & 2 & 9 & 7 & 111 & \textbf{13.42} & 29.41 \\
			AdvUnlearn\cite{zhang2024defensive} & 12 & 7 & 4 & 13 & 6 & 2 & 0 & 8 & 52 & 15.35 & 29.3 \\
			RealEra\cite{liu2024realera} & 19 & 6 & 2 & 37 & 23 & 4 & 0 & 2 & 93 & - & - \\
			SPEED \cite{li2025speed}  & 20 & 42 & 7 & 3 & 29 & 2 & 5 & 5 & 113 & 37.82 & 26.29 \\
			SalUn~\cite{fan2023salun} & 2 & 14 & {0} & 14 & {7} & 2 & 7 & 5 & 51 & -- & -- \\
			CE-SDWV \cite{tu2025sdwv} & 13 & 46 & 2 & {2} & 13 & 0 & 1 & 6 & 84 & 13.66 & 30.80 \\
			SPM \cite{lyu2024one} & 22 & {4} & 9 & 12 & 4 & 0 & 0 & 5 & 56 & - & - \\
			RECE \cite{gong2024reliable} & 17 & 23 & 0 & 8 & 8 & 0 & 6 & 4 & 66 & - & - \\ 
			SDD \cite{kim2023towards} & 14 & {4} & 7 & 3 & 8 & 1 & 0 & 4 & 41 & - & - \\
			DuMo \cite{han2025dumo} & 8 & 6 & 2 & 7 & {1} & 4 & 0 & 6 & 34 & - & - \\
			ACE \cite{wang2025ace} & 5 & 7 & 3 & 6 & 2 & 3 & 4 & 9 & 39 & 14.69 & 30.80 \\
%			\midrule
			Ours~ & {1} & {5} & 2 & {4} & 8 & 2 & {0} & 1 & \textbf{23} & 14.44 & 30.64 \\
			\midrule
			SD v1.4 & 148 & 170 & 29 & 63 & 266 & 18 & 42 & 7 & 743 & 14.04 & 31.34 \\
			SD v2.1 & 105 & 159 & 17 & 60 & 177 &  9 & 57 & 2 & 586 & 14.87 & 31.53 \\
			\bottomrule
		\end{tabular}%
	}
	%		\vspace{-0.2cm}
	\label{tab:nudity}
\end{table*}

In this section, we present a comprehensive evaluation of our proposed method by benchmarking it against SOTA baselines on both single-concept erasure (NSFW removal; Section~\ref{exp:nsfw}) and multi-concept erasure tasks, including 100-celebrity erasure (Section~\ref{exp:cele}) and 100-artistic style erasure (Section~\ref{exp:art}). 
%The comparison includes the following baselines: ESD-u \cite{gandikota2023erasing}, ESD-x \cite{gandikota2023erasing}, FMN \cite{zhang2023forget}, SLD-M \cite{schramowski2023safe}, UCE \cite{gandikota2023unified}, and AC \cite{kumari2023ablating}. 
Finally, we perform ablation studies (Section~\ref{sec:ablation}) to assess the contribution of key components in our approach.

\subsection{Implementation Details}
We finetune all models based on Stable Diffusion (SD) v1.4 and generate outputs using the DDIM sampler~\cite{song2020denoising} over 50 inference steps. Our experimental setup follows the settings described in MACE~\cite{lu2024mace}. Each LoRA module undergoes 50 gradient update steps during training. For the baselines, we adopt the configurations provided in their respective original implementations. 

\subsection{Erasing NSFW Content}
\label{exp:nsfw}

\noindent\textbf{Configuration.} In this experiment, we focus on removing the concept ``nudity'' from the model, representing a typical NSFW category. Specifically, we follow the ``nudity'', ``naked'', ``erotic'', ``sexual'' prompts introduced in~\cite{heng2023selective,lu2024mace} to guide the construction of the concept-specific saliency map $\boldsymbol{M}^*$ over the UNet. Based on $\boldsymbol{M}^*$, we finetune SD~v1.4 to eliminate the concept. 

For evaluation, we use the full set of 4,703 prompts from the I2P dataset~\cite{schramowski2023safe} along with their corresponding random seeds to generate images. We then apply NudeNet~\cite{nudenet} with the threshold of 0.6 to detect exposed body parts in the sampled images, treating the detection results as an indicator of residual nudity in the model's output. In addition, we assess the effectiveness of concept removal techniques in preserving benign content, utilizing the MS-COCO dataset~\cite{lin2014microsoft}. We sample 30,000 captions from the validation split to generate images and compute FID~\cite{parmar2022aliased} and CLIP score~\cite{radford2021learning} as metrics for image quality and semantic alignment.

%\vspace{0.1cm}
\noindent\textbf{Results Analysis.} The experimental results are presented in Table~\ref{tab:nudity}. Our method generates significantly less NSFW content under the I2P benchmark prompts compared to other baselines, especially in challenging regions such as breasts. At the same time, our method also achieves competitive performance in terms of FID and CLIP scores. These results demonstrate that our approach can effectively remove explicit content from the model without compromising image quality.

\subsection{Erasing Celebrity} 
\label{exp:cele}

\begin{table}[tbp]
	\centering
	\vspace{0.07cm}
	\caption{Results of Erasing Celebrity. We report the accuracy for erased celebrities ($\text{Acc}_e$), accuracy for preserved celebrities ($\text{Acc}_p$), harmonic mean metric ($H_c$) and the proportion of clearly recognizable faces (Face Ratio). FID and CLIP are results based on MS-COCO dataset. SD v1.4 and SD v2.1 are used as reference base models.}
	\vspace{-0.2cm}
	\resizebox{\columnwidth}{!}{
		\begin{tabular}{lcc>{\columncolor{mypink}}cccc}
			\toprule
			Method & $\text{Acc}_e \downarrow$ & ${\text{Acc}_p \uparrow}$ & ${H_c \uparrow}$ & {Face Ratio}$\uparrow$ & {FID}$\downarrow$ & {CLIP}$\uparrow$ \\
			\midrule
			FMN~\cite{zhang2023forget} & 0.9223 & \textbf{0.9076} & 0.1431 & \textbf{0.9940} & 13.95 & \textbf{31.31} \\
			ESD-x~\cite{gandikota2023erasing} & 0.2784 & 0.2793 & 0.4027 & 0.8088 & 14.65 & 28.90 \\
			ESD-u~\cite{gandikota2023erasing} & \textbf{0.0406} & 0.3909 & 0.4598 & 0.4724 & 15.14 & 29.02 \\
			SLD-M~\cite{schramowski2023safe} & 0.8706 & 0.7946 & 0.2237 & 0.9093 & 17.54 & 30.93 \\
			AC~\cite{kumari2023ablating} & 0.8913 & 0.9096 & 0.1977 & 0.9932 & 13.92 & 31.23 \\
			UCE~\cite{gandikota2024unified}   & 0.0012 & 0.3790 & 0.5495 & 0.7179 & 106.57 & 19.17 \\
			RECE \cite{gong2024reliable}       & 0.0243 & 0.2371 & 0.3816 & -      & 177.57 & 12.09 \\
			SPEED \cite{li2025speed}     & 0.0587 & 0.8554 & 0.8963 & -      & 44.97 & 26.22 \\
			MACE~\cite{lu2024mace} & 0.0430 & 0.8456 & 0.8979 & 0.9820 & 12.82 & 30.21 \\
			Ours       & 0.0430 & 0.8807 & \textbf{0.9173} & 0.9816 & \textbf{11.71} & 30.40 \\
			\midrule
			SD v1.4 & 0.9648 & 0.9388 & - & 0.9876 & 14.04 & 31.34 \\
			SD v2.1 & 0.9324 & 0.9293 & - & 0.9879 & 14.87 & 31.53 \\
			\bottomrule
	\end{tabular}}
	\label{tab:celebrity}
\end{table}

\begin{figure*}[tbp]
	\centering
	%		\vspace{-0.2cm}
	\includegraphics[width=\linewidth]{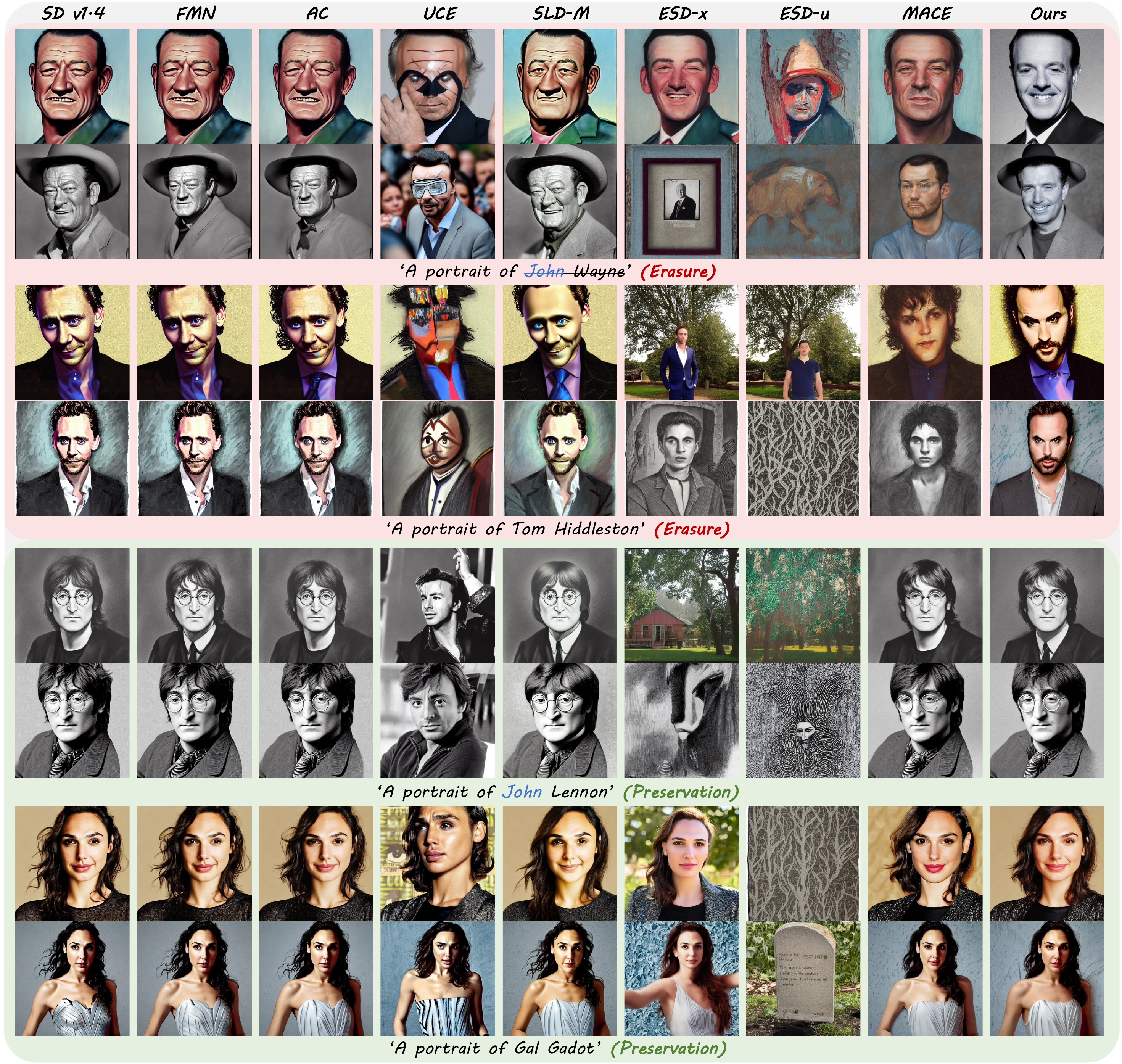}
%	\vspace{-0.3cm}
	\caption{Qualitative comparison of erasing 100 celebrities from SD v1.4. John Wayne and Tom Hiddleston are in the erasure group for evaluating erasure performance; John Lennon and Gal Gadot are in preservation group for assessing preservation performance. Preserving John Lennon is challenging due to the shared first name with John Wayne.}
	%		\vspace{-0.3cm}
	\label{fig:cele}
\end{figure*}

\noindent\textbf{Configuration.} In this section, we evaluate the performance of our method on the task of simultaneously erasing multiple celebrity concepts, using the 200-celebrity dataset from MACE~\cite{lu2024mace}, which includes 100 celebrity concepts designated for erasure and 100 concepts intended to be preserved.

We conduct experiments by finetuning SD~v1.4 to erase all 100 celebrity identities in the erasure group. We evaluate the effectiveness of our method by generating portraits of the targeted celebrities. Successful erasure is indicated by a low top-1 accuracy from GIPHY Celebrity Detector (GCD)~\cite{GCD} in identifying the erased identities. Additionally, to investigate the influence of our method on celebrities in the preservation group, we generate and evaluate their portraits in the same manner, where a high top-1 GCD accuracy reflects minimal impact on these preserved identities. We also report the harmonic mean \(H_c\) metric introduced in~\cite{lu2024mace}, which provides a balanced evaluation of the trade-off between successful erasure of unwanted celebrity concepts and the preservation of unrelated ones:
\begin{equation}
	H_c = \frac{1}{(1-\text{Acc}_e)^{-1} + (\text{Acc}_p)^{-1}},
\end{equation}
where $H_c$ is the harmonic mean for celebrity erasure, $\text{Acc}_e$ is the accuracy for the erased celebrities, and $\text{Acc}_p$ for the preserved ones.

\vspace{0.1cm}
\noindent\textbf{Results Analysis.} Figure~\ref{fig:cele} shows the qualitative comparison. Table~\ref{tab:celebrity} summarizes the performance of baselines on the celebrity concept erasure task. Our method achieves the highest $H_c$, outperforming all baselines and highlighting an excellent balance between concept erasure and preservation of unrelated ones.

Our method obtains the lowest FID score, surpassing all compared baselines and even the original SD models. A plausible reason for this improvement is that our finetuning process, while primarily intended for erasing specific concepts, implicitly regularizes the model by encouraging more consistent representations of general concepts. Additionally, the CLIP score remains competitive, indicating minimal disruption to semantic alignment.

\subsection{Erasing Art Style}
\label{exp:art}
\noindent\textbf{Configuration.} For art style erasure, we follow a similar training procedure as described in Section~\ref{exp:cele}, with certain hyperparameter adjustments detailed in the Appendix. To evaluate performance, we use the 200-artist dataset from MACE~\cite{lu2024mace}, which consists of two groups: an erasure group of 100 artists whose styles are targeted for removal, and a preservation group of 100 artists whose styles are intended to be retained. 

We use the CLIP score to assess how well the generated images align with the intended artistic style. For the erasure group, a lower CLIP score ($\text{CLIP}_e$) indicates better performance, as it suggests more effective removal of the target concept. In contrast, for the preservation group, a higher CLIP score ($\text{CLIP}_p$) is desirable, as it reflects minimal disruption to unrelated concepts. The overall performance is captured by $H_a = \text{CLIP}_p - \text{CLIP}_e$, where a higher value indicates better balance between preservation and erasure.

\vspace{0.1cm}
\noindent\textbf{Results Analysis.} Table \ref{tab:art} summarizes the performance of our method in erasing artistic styles. Our method achieves the highest \(H_a\), substantially surpassing all baseline methods, demonstrating superior balance in effectively removing targeted art styles and preserving unrelated art styles. Considering the overall performance across other metrics, our strategy shows notable competitiveness compared to existing approaches. 

\begin{table}[tbp]
	\centering
	\caption{Results of Erasing 100 Art Styles. We report the CLIP score for erased artistic style ($\text{CLIP}_e$), CLIP score for preserved artistic style ($\text{CLIP}_p$), the overall score ($H_a$). FID and CLIP are results based on MS-COCO dataset.}
%	\vspace{-0.3cm}
	\resizebox{\linewidth}{!}{
		\begin{tabular}{lcc>{\columncolor{mypink}}ccc}
			\toprule
			{Method} & $\text{CLIP}_e\downarrow$ & $\text{CLIP}_p\uparrow$ & $H_a\uparrow$ & FID-COCO$\downarrow$ & CLIP-COCO$\uparrow$ \\ 
			\midrule
			FMN~\cite{zhang2023forget}   & 29.63 & \textbf{28.90} & -0.73 & 13.99 & \textbf{31.31} \\
			ESD-x~\cite{gandikota2023erasing}  & 20.89 & 21.21 & 0.32 & 15.19 & 29.52 \\
			ESD-u~\cite{gandikota2023erasing} & \textbf{19.66} & 19.55 & -0.11 & 17.07 & 27.76 \\
			SLD-M~\cite{schramowski2023safe} & 28.49 & 27.89 & -0.60 & 17.95 & 30.87 \\
			AC~\cite{kumari2023ablating} & 29.26 & 28.54 & -0.72 & 14.08 & 31.29 \\
			UCE~\cite{gandikota2024unified} & 21.31 & 25.70 & 4.39 & 77.72 & 19.17 \\
			MACE~\cite{lu2024mace} & 22.59 & 28.58 & 5.99 & \textbf{12.71} & 29.51 \\
			Ours & 20.6 & 26.78 & \textbf{6.18} & 12.96 & 27.63 \\
			\midrule
			SD v1.4 & 29.63 & 28.90 & - & 14.04 & 31.34 \\
			\bottomrule
	\end{tabular}}
	\label{tab:art}
\end{table}

\subsection{Ablation Study}
\label{sec:ablation}
To investigate the contribution of key components in our approach, we conduct ablation studies on both multiple concepts (celebrity removal) and single concept (NSFW removal) tasks. The experimental configurations and corresponding results are presented in Tables~\ref{tab:ablation_cele} and \ref{tab:ablation_nsfw}, respectively. 

We begin by ablating each component of our loss function in the context of celebrity removal. \textbf{Config~A}, which applies $\mathcal{L}_{\text{erase}}$ across all stages without preserving the early-stage score function field, shows strong removal capability but clearly suffers in terms of preservation. \textbf{Config~B} builds on Config~A by adding $\mathcal{L}_{\text{uncond}}$ to maintain the unconditional score function, resulting in improved overall performance in terms of $H_c$. Next, \textbf{Config~C} applies $\mathcal{L}_{\text{erase}}$ only during the mid-to-late sampling stages, aiming to avoid disruption of the early-stage score function field. While this slightly improves $H_c$, the results remain unsatisfactory. \textbf{Config~D} enhances Config~C by adding $\mathcal{L}_{\text{uncond}}$, applied over the same timesteps as $\mathcal{L}_{\text{erase}}$, which further improves overall performance. \textbf{Config~E} extends Config~C by introducing $\mathcal{L}_{\text{preserve}}$, which helps retain the original score function field and significantly boosts preservation performance in terms of $\text{Acc}_p$. Our full method builds upon Config~E by applying $\mathcal{L}_{\text{uncond}}$ across all stages, resulting in superior overall performance.

\begin{table}[tbp]
	\centering
	\caption{Ablation study on multiple concepts (celebrity) removal. $\mathcal{L}_\text{erase}^*$: $\mathcal{L}_\text{erase}$ is applied at all denoising timesteps during training. $\mathcal{L}_\text{erase}$: $\mathcal{L}_\text{erase}$ is applied only during the mid-to-late stages of the denoising process in training.}
%	\vspace{-0.3cm}
	\resizebox{\linewidth}{!}{
		\begin{tabular}{c ccccc cc>{\columncolor{mypink}}c}
			\toprule
			\multirow{2}{*}{{Config}} & \multicolumn{5}{c}{{Components}} & \multicolumn{3}{c}{{Metrics}} \\
			\cmidrule(lr){2-6} \cmidrule(lr){7-9}
			& $\mathcal{L}_\text{erase}$ & $\mathcal{L}_\text{erase}^*$& $\mathcal{L}_\text{preserve}$ & $\mathcal{L}_\text{uncond-early}$ & $\mathcal{L}_\text{uncond-late}$ & $ \text{Acc}_e \downarrow$ & $ \text{Acc}_p \uparrow$ & $H_c \uparrow$ \\
			\midrule
			A   & \faTimes & \faCheck & \faTimes & \faTimes & \faTimes & 0.0192 & 0.7785 & 0.8680 \\
			B   & \faTimes & \faCheck & \faTimes & \faCheck & \faCheck & \textbf{0.0042} & 0.7848 & 0.8778 \\
			C   & \faCheck & \faTimes & \faTimes & \faTimes & \faTimes  & 0.0309 & 0.8094 & 0.8821 \\
			D   & \faCheck & \faTimes & \faTimes & \faTimes & \faCheck  & 0.0075 & 0.8013 & 0.8867 \\
			E   & \faCheck & \faTimes & \faCheck & \faTimes & \faTimes  & 0.0910 & 0.8545 & 0.8809 \\
			%			F   & \faCheck & \faTimes & \faCheck & \faCheck & \faTimes & 0.6216 & 0.5880 & 0.4605 \\
			\midrule
			{Ours} &  & \faTimes & \faCheck & \faCheck & \faCheck  & 0.0430 & \textbf{0.8807}  & \textbf{0.9173} \\
			%			\midrule
			%			SD v1.4     & -         & -         & -       & -     & -    & 0.9648 & 0.9388 & 0.0679 \\
			\bottomrule
		\end{tabular}
	}	
	%	\vspace{-0.2cm}
	\label{tab:ablation_cele}
\end{table}

For NSFW content removal, \textbf{Config~F} finetunes the entire UNet, while \textbf{Config~G} finetunes only a subset of parameters using a saliency map obtained from a single calculation. However, Config~G performs worse than Config~F, suggesting that a saliency map generated from a single pass may be inaccurate. In contrast, our method derives a more precise concept-specific saliency map by taking the intersection of multiple saliency maps calculated from different prompts and seeds. This allows us to more accurately identify the parameters strongly associated with the concept, leading to substantially improved performance.

\begin{table}[tbp]
	\centering
	\caption{Ablation study on single concept (NSFW) removal. Single Map: $\boldsymbol{M}^*$ is generated using a single prompt and one random seed. Multi Maps: $\boldsymbol{M}^*$ is generated taking the intersection of saliency maps obtained using multiple prompts and multiple random seeds. }
%	\vspace{-0.3cm}
	\resizebox{\linewidth}{!}{
		\begin{tabular}{c cc ccc>{\columncolor{mypink}}c}
			\toprule
			\multirow{2}{*}{{Config}} & \multicolumn{2}{c}{{Components}} & \multicolumn{4}{c}{{Inappropriate Image Prompt (I2P)}} \\
			\cmidrule(lr){2-3} \cmidrule(lr){4-7}
			& {Single Map} & {Multi Maps} & Breasts (M\&F) & Genitalia (M\&F) & Others & Total$\downarrow$\\
			\midrule
			F  & \faTimes & \faTimes  & 136 & 11 & 148 & 295 \\
			G   & \faCheck & \faTimes  & 83 & 56 & 184 & 323 \\
			\midrule
			{Ours} & \faTimes & \faCheck & \textbf{8} & \textbf{3} & \textbf{12} & \textbf{23} \\
			%					\midrule
			%					SD v1.4     & -         & -   & 308 & 25 & 410 & 743\\
			\bottomrule
		\end{tabular}
	}
	%		\vspace{-0.2cm}
	\label{tab:ablation_nsfw}
\end{table}

\section{Conclusion}
Our work introduces a geometric perspective on concept erasure within diffusion models. Utilizing this perspective, we found that reversing the condition direction of classifier-free guidance during the mid-to-late stages of the denoising process allows for modifying detailed content without compromising the overall structural integrity of the generated images. Inspired by this insight, we propose ANT, a novel framework that effectively balances the removal of unwanted concepts while preserving unrelated elements. ANT demonstrates superior performance in both single- and multi-concept erasure scenarios, significantly outperforming current SOTA methods.

{
    \small
    \bibliographystyle{ieeenat_fullname}
    \bibliography{main}
}

\clearpage
%\onecolumn
\appendix

\noindent\textbf{\Large Appendix}
\renewcommand\thesection{\Alph{section}}
\setcounter{section}{0}

\section{Additional Related Work}
With the advancement of deep learning~\cite{li2024towardsa,li2024distinct,li2024towardsb,yu2025prnet,yu2024scnet,yu2024ichpro,yang2025llm,yang2025wastewater,yang2025dual,yang2024harnessing,khan2025baiot,li2022vpai_lab,li2022multi} and generative models~\cite{chang2023muse, ding2022cogview2, nichol2021glide, ramesh2022hierarchical, rombach2022high, saharia2022photorealistic,lu2023tf,lu2024robust,zhu2024oftsr}, an increasing number of studies have begun to focus on the issue of concept erasure in generative models.

\subsection{Balancing Erasure and Preservation.} With the advancement of concept erasure techniques, the community has come to recognize that concept erasure should not only focus on the target concept but also aim to minimize the impact on unrelated concepts during finetuning. Numerous studies~\cite{lyu2024one,wang2025ace,wang2025ace2,kim2023towards,kim2024safeguard,bui2024erasing,bui2025fantastic,gaintseva2025casteer,gao2024meta,chavhan2024conceptprune,heng2023selective,han2025dumo,wu2024munba,liu2024realera,lu2024mace,thakral2025fine,han2024continuous,huang2024receler,yang2024pruning,shirkavand2024efficient,chengrowth,schioppa2024model,zhao2024advanchor,tu2025sdwv,fuchi2025erasing,meng2025concept,tian2025sparse,xue2025crce,thakral2025continual,li2025detect,chen2025safe,gao2024eraseanything} emphasize the model’s balanced performance between the target concept and unrelated concepts. MACE~\cite{lu2024mace} introduces concept-focal importance sampling and modular LoRA integration, allowing for scalable multi-concept erasure while avoiding interference across modules. Several works~\cite{liu2024realera,xue2025crce} explore semantic-aware preservation by modeling relationships between erased and retained concepts, improving quality retention in adjacent concept spaces. Some frameworks~\cite{wu2024munba,thakral2025fine} formalize the forgetting–retention trade-off, offering principled mechanisms to control degradation. 

\subsection{Finetuning Efficiency.} In addition to balancing erasure and preservation, several recent methods~\cite{lyu2024one,wang2025ace2,gong2024reliable,li2025speed,kim2025concept,fuchi2024erasing,bui2024erasing,gandikota2024unified,meng2025concept} have increasingly emphasized einetuning efficiency to meet practical demands. \cite{gandikota2024unified,li2025speed} achieve erasure across hundreds of concepts within seconds by leveraging low-rank adapters or null-space constraints, enabling rapid adaptation across diffusion model variants. \cite{gong2024reliable,wang2025ace2} introduce closed-form or structure-aware updates that reduce erasure time by orders of magnitude. These advancements demonstrate a trend toward minimal-latency, high-throughput concept erasure that enables practical integration into production-scale text-to-image pipelines.

\subsection{Scalability.} With the growing demand for safe and policy-compliant generative models, scalable multi-concept erasure techniques\cite{xiong2024editing,kim2023towards,chen2025trce,li2025speed,lu2024mace,lee2025localized,gandikota2024unified,chen2025safe} have emerged as a key direction in diffusion model editing. \cite{gandikota2024unified} introduces an editing framework that supports the simultaneous modification of multiple concepts through lightweight model updates. \cite{lu2024mace} leverages modular LoRA-based editing combined with closed-form integration to eliminate over 100 concepts with minimal interference. \cite{xiong2024editing} adopts a two-stage process involving self-distillation and multi-layer editing, scaling up to 1,000 concepts while preserving specificity and visual fidelity. Additional methods such as \cite{chen2025trce,chen2025safe} enhance scalability through embedding-space operations or adversarially robust training objectives. These techniques collectively push concept erasure toward broader, more practical deployment scenarios requiring high-volume, reliable editing.

\subsection{Robustness.} Despite successful concept suppression, erased models remain vulnerable to adversarial prompts that can reactivate undesirable content. A growing number of methods\cite{srivatsan2024stereo,wang2025ace,gong2024reliable,jain2024trasce,beerens2025vulnerability,chen2025trce,gaintseva2025casteer,kim2025concept,zhang2024defensive,chavhan2024conceptprune,gao2024meta,nguyen2024unveiling,wu2024munba,lee2025localized,park2024direct,huang2024receler,kim2024race,yang2024pruning,tu2025sdwv,cywinski2025saeuron,meng2025concept,hu2025safetext,tian2025sparse,li2025detect} have begun to address this issue explicitly, aiming to improve model reliability in the face of prompt-based attacks. Methods such as \cite{srivatsan2024stereo,jain2024trasce} tackle this by pairing adversarial prompt discovery with robust erasure objectives or inference-time steering, offering stronger defense without retraining. Others, like \cite{zhang2024defensive,park2024direct,kim2024race} incorporate adversarial training or preference-based optimization directly into the unlearning process to improve stability against attack. Complementary strategies from \cite{nguyen2024unveiling,cywinski2025saeuron,hu2025safetext} focus on interpretable attribution or encoder-level alignment to neutralize unsafe inputs at their origin. Together, these works underscore the need for erasure techniques that are not only effective but resilient under adversarial conditions.

\section{Hyperparameters Setup}
Table \ref{tab:erasure_hyperparams} presents the specific hyperparameters used in the experiments for erasing different types of concepts. 
\begin{table}[tbp]
	\centering
	\caption{Training hyperparameters for NFSW content, celebrity and art style erasure tasks.}
	\vspace{-0.3cm}
	\resizebox{\columnwidth}{!}{
		\begin{tabular}{ccccccc}
			\toprule
			\textbf{Erasure Type}  & \textbf{Learning Rate} & \textbf{Epochs} & $\lambda_1$ & $\lambda_2$ & $\lambda_3$ & $t'$ \\
			\midrule
			NFSW Content & $5.0 \times 10^{-4}$ & 250 & 1.0 & 0.5 & 0.5 & 43 \\
			\midrule
			Celebrity & $5.0 \times 10^{-4}$ & 400 & 0.4 & 0.5 & 0.2 & 40 \\
			\midrule
			Art Style & $5.0 \times 10^{-4}$ & 400 & 0.4 & 0.5 & 0.2 & 47 \\
			\bottomrule
		\end{tabular}
	}
%	\vspace{-0.03cm}
	\label{tab:erasure_hyperparams}
\end{table}

\section{Limitations and Future Work}

Our work has primarily been tested on UNet-based diffusion models~\cite{sd14modelcard,stable2.1}. As diffusion models increasingly adopt architectures like MMDiT~\cite{peebles2023scalable,esser2024scaling,flux}, evaluating the compatibility of our approach with these new frameworks will be a key focus of our next phase. Additionally, assessing the robustness of our framework against adversarial prompts~\cite{tsai2023ring,zhang2024generate} and its ability to withstand methods for learning personalized concepts~\cite{gal2022image,ruiz2023dreambooth} will be of critical importance.

\section{Additional Qualitative Results}
\begin{figure*}[t]
	\centering
	\includegraphics[width=\textwidth]{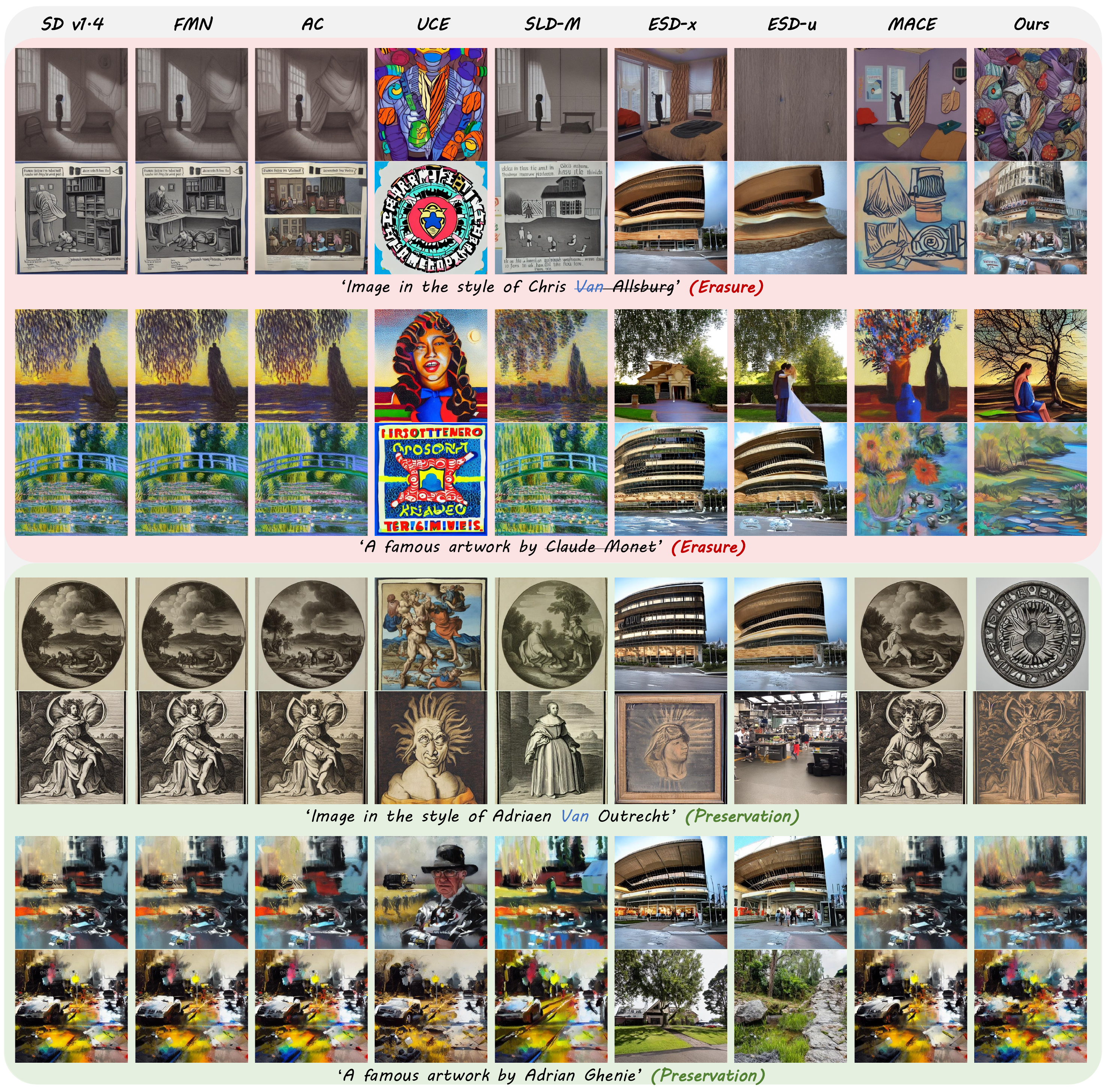}
	%	\vspace{-1cm}
	\caption{Qualitative comparison on art style erasure. The images on the same row are generated using the same random seed. Chris Van Allsburg and Claude Monet are in the erasure group, while Adriaen Van Outrecht and Adrian Ghenie are in the retention group.}
	\label{fig:fullpage}
\end{figure*}

Figure \ref{fig:fullpage} presents a qualitative comparison of art style erasure and the preservation of unrelated concepts across different baselines. In the erasure rows, our approach effectively eliminates the target artistic styles (Chris Van Allsburg and Claude Monet) while retaining high-quality, plausible generation. In the preservation rows, our method successfully maintains the visual characteristics of unrelated artists (Adriaen Van Utrecht and Adrian Ghenie), showing minimal unintended impact on non-target styles.

\end{document}